\documentclass[twoside,11pt]{article}

\usepackage{jmlr2e}
\usepackage{jlcode}

\jmlrheading{1}{2000}{1-48}{4/00}{10/00}{David Muhr and Michael Affenzeller}

\ShortHeadings{OutlierDetection.jl ecosystem for Julia}{Muhr, Affenzeller
and Blaom.}
\firstpageno{1}

\begin{document}

\title{OutlierDetection.jl: A modular outlier detection ecosystem for the Julia programming language}

\author{\name David Muhr\textsuperscript{1,2} \email david.muhr@bmw.com \\
  \name Michael Affenzeller\textsuperscript{2,3} \email
  michael.affenzeller@heuristiclab.at \\
  \name Anthony D.
  Blaom\textsuperscript{4} \email a.blaom@auckland.ac.nz \\ \addr{
    \textsuperscript{1} BMW Group, Steyr, Austria.
    \\
    \textsuperscript{2}
    Institute for Formal Models and Verification, Johannes Kepler University Linz,
    Austria.
    \\ \textsuperscript{3}
    HEAL, University of Applied Sciences Upper Austria, Austria.
    \\ \textsuperscript{3}
    University of Auckland, New Zealand.
  }}


\maketitle

\begin{abstract}
  OutlierDetection.jl is an open-source ecosystem for outlier detection in Julia.
  It provides a range of high-performance outlier detection algorithms
  implemented directly in Julia.
  In contrast to previous packages, our ecosystem enables the development
  highly-scalable outlier detection algorithms using a high-level programming
  language.
  Additionally, it provides a standardized, yet flexible, interface for future
  outlier detection algorithms and allows for model composition unseen in
  previous packages.
  Best practices such as unit testing, continuous integration, and code coverage
  reporting are enforced across the ecosystem.
  The most recent version of OutlierDetection.jl is available at
  \url{https://github.com/OutlierDetectionJL/OutlierDetection.jl}.
\end{abstract}


\section{Introduction}

An outlier is "an observation (or subset of observations) which appears to be
inconsistent with the remainder of that set of data"
\citep{barnettOutliersStatisticalData1978}.
Outlier detection is the research area that studies the detection of such
inconsistent observations.
Outliers are by nature infrequent events, thus labels are difficult to obtain
and the ground truth is often absent in outlier detection tasks.
Outlier detection is mainly used in fields that process large amounts of
unlabelled data, such as network intrusion detection, fraud detection, medical
diagnostics or industrial quality control.
For reviews covering the most important outlier detection application areas
refer to \cite{chandolaAnomalyDetection2009} and
\cite{pimentelReviewNoveltyDetection2014}.

Software packages for outlier detection exist in various programming languages
such as \textsl{ELKI Data Mining} \citep{achtertVisualEvaluationOutlier2010} in
Java, \textsl{DDoutlier} \citep{madsenDDoutlierDistanceDensityBased2018} in R,
or \textsl{PyOD} \citep{zhaoPyODPythonToolbox2019} in Python.
Existing outlier detection packages, however, either cater to a research
community, where benchmark datasets are small and performance is negligible, or
are plagued by the "two language problem" to define algorithms.
The two language problem refers to the necessary usage of multiple programming
languages to achieve high-performing numerical code, e.g., Python and NumPy,
calling C functions under the hood \cite{bezansonJuliaFreshApproach2017}.
\textsl{OutlierDetection.jl} aims to address these problems, allowing future
researchers to define scalable algorithms using the Julia programming language
\footnote{\url{https://julialang.org}}.

\section{Design Goals}

The following points summarize the key design goals we followed in the
development of \textsl{OutlierDetection.jl}.
Most importantly, we choose to contribute to Julia's existing machine learning
community instead of building yet another separate machine learning package.

\begin{description}
  \item[Community.] An open-source community
    \footnote{\url{https://github.com/OutlierDetectionJL}}
    has been founded to facilitate the collaboration of ecosystem contributors.
    Additionally, tight integration to the rest of Julia's machine learning
    ecosystem is provided, integrating with the most common machine learning
    libraries and directly extending Machine Learning in Julia (MLJ)
    \citep{blaomMLJJuliaPackage2020}.
  \item[Modularity.] The functionality is split into multiple packages, and
    each package fulfills a specific purpose, see Figure \ref{architecture}
    for an overview.
    A modular package structure has shown to be beneficial in other large Julia
    projects such as \textsl{POMDPs.jl} \citep{egorovPOMDPsJlFramework2017} or
    \textsl{MLJ} \citep{blaomMLJJuliaPackage2020}.
    On the one hand, this modularity enables developers to contribute more easily,
    and, on the other hand, the complexity is hidden from the end-users, which have
    a single point of entry that does not load all dependencies upfront,
    which is especially important with Julia's precompilation.
  \item[Quality.] Unit testing and code coverage reporting is used to test the
    internal procedures of all packages in the ecosystem.
    Continuous integration is used to conduct automated testing under various
    versions of Julia and operating systems.
    After each commit, or when a pull request is opened, automatic cross-platform
    tests are executed to ensure that the code quality meets our standards.
  \item[Documentation.] Documentation is developed in a unified notion for the
    developers and users of \textsl{OutlierDetection.jl} by
    standardizing the comment structure and directly transforming comments to
    user-readable documentation of the entire API using \textsl{Documenter.jl}
    \footnote{\url{https://github.com/JuliaDocs/Documenter.jl}}.
    Additional documentation such as usage examples are written in markdown with
    executable code blocks, ensuring that the documentation does not get out of
    sync with the codebase.
  \item[Relevance.] \textsl{OutlierDetection.jl} is the first major outlier
    detection project for the Julia programming language.
    It is the first project that tackles the two-language problem in outlier detection,
    enabling researchers to develop high-performing outlier detection algorithms that scale
    to millions of data points with comparably little effort.
  \item[Standardization.] Typically, each novel outlier detection algorithm
    comes with its own set of assumptions, for example, that the outlier class
    is defined as \textsc{1} or \textsc{-1}.
    Using a new concept called \textit{scientific types}
    \citep{kiralyDesigningMachineLearning2021}, we can standardize such decisions
    without limiting the algorithm authors' flexibility in their machine-type
    representation.
\end{description}

\section{Implementation}

As mentioned previously, one of the goals of this library is to integrate with
the rest of Julia's machine learning ecosystem.
Julia's multiple dispatch \citep{bezansonJuliaFreshApproach2017} enables us to
augment other libraries and compose functionality across packages and
organizations.
One example of this composability is the integration with MLJ, which we use as
an entry point for users to discover outlier detection algorithms.

\begin{figure}[h]
  \centering
  \includegraphics[width=\textwidth]{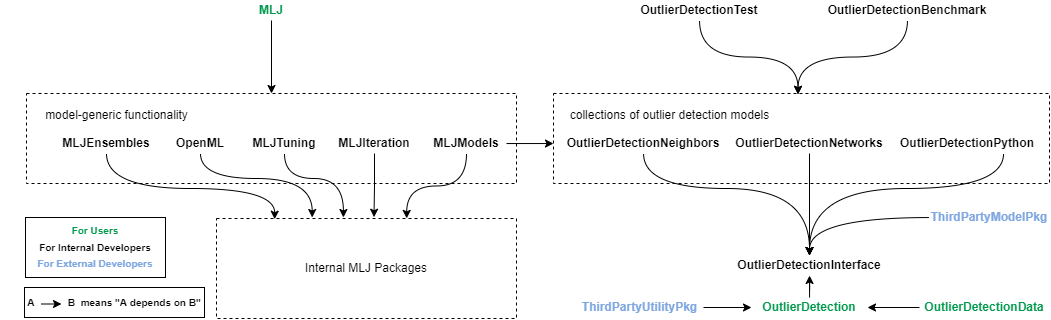}
  \caption{\textsl{OutlierDetection.jl} package overview and MLJ integration}
  \label{architecture}
\end{figure}

In Figure \ref{architecture}, we show the conceptual architecture of the
packages contained in the outlier detection organization and how they
interoperate with MLJ.
All outlier detection models are directly usable in MLJ, without knowing any of
the underlying package names beforehand.
This is possible because all models are added to MLJ's model registry.
Listing all unsupervised outlier detection models is shown in Figure
\ref{list-models}.

\begin{figure}[h]
  \jlinputlisting{code/list-models.jl}
  \caption{Listing all unsupervised outlier detection models.}
  \label{list-models}
\end{figure}

One of the major design decisions is to separate the tasks occuring in outlier
detection into separate packages.
Outlier detection algorithms typically assign an outlier score to each instance
in the dataset \citep{aggarwalOutlierAnalysis2017}, as visible in Figure
\ref{transform}.

\begin{figure}[h!]
  \jlinputlisting{code/transform.jl}
  \caption{Transform a number of points into raw outlier scores.}
  \label{transform}
\end{figure}

The conversion of outlier scores to labels or probabilities is a separate task
usually based on a threshold identified from the training data.
Because scoring and score conversion are mostly separate tasks, we centralize
various score conversion utilities in the \textsl{OutlierDetection.jl} package.
Therefore, developers of new detection algorithms do not have to implement
score conversion utilities themselves but are free to implement their utilities
if required.
We provide multiple ways to convert scores; one of the approaches is using a
model wrapper as shown in Figure \ref{predict}.
There is one more user-facing package, not yet mentioned in the examples,
namely \textsl{OutlierDetectionData.jl}.
This package provides a unified interface to load common outlier detection benchmark
datasets as shown in Figure \ref{load-data}.

\begin{figure}[h!]
  \jlinputlisting{code/predict.jl}
  \caption{Predict labels instead from the raw outlier scores.}
  \label{predict}
\end{figure}

\begin{figure}[h!]
  \jlinputlisting{code/load-data.jl}
  \caption{Load a dataset from the ODDS collection
    \citep{rayanaODDSLibrary2016}.}
  \label{load-data}
\end{figure}

The integration with MLJ enables sophisticated use cases to be defined in a
declarative fashion.
For example, hyperparameter tuning, model evaluation, ensemble learning, or
complex model composition in terms of learning networks can be achieved
\citep{blaomFlexibleModelComposition2020}.
Ensemble techniques have shown to be beneficial in many outlier detection tasks
\citep{aggarwalTheoreticalFoundationsAlgorithms2015} and model composition is
recently investigated, for example combining deep self-supervised
representation learning with nearest-neighbor search
\citep{bergmanDeepNearestNeighbor2020}.

\section{Conclusion}

This paper presents an outlier detection ecosystem for Julia that allows
researchers to implement scalable outlier detection algorithms in a high-level
programming language.
A carefully designed library interface enables model composition unseen in
previous outlier detection packages.
One key factor enabling model composition is the clear separation of scoring
and score conversion of outlier detection models.
We believe that OutlierDetection.jl will allow future researchers to quickly
iterate on new outlier detection algorithms, and model composition will enable
novel outlier detection use cases and methods.

\bibliography{references.bib}

\end{document}